\documentclass{article}

\usepackage{microtype}
\usepackage{graphicx}
\usepackage{subfigure}
\usepackage{enumitem}
\setlist[1]{itemsep=-5pt}
\usepackage{booktabs} 
\usepackage[ruled,vlined,algo2e]{algorithm2e}

\SetCommentSty{mycommfont}

\SetKwInput{KwInput}{Input}                
\SetKwInput{KwOutput}{Output}              

\usepackage{hyperref}

\usepackage[accepted]{icml2019}

\icmltitlerunning{Learning RTMs for Climate Change Applications}

\begin{document}

\twocolumn[
\icmltitle{Learning Radiative Transfer Models for Climate Change Applications in Imaging Spectroscopy}




\begin{icmlauthorlist}
\icmlauthor{Shubhankar Deshpande}{cmuri}
\icmlauthor{Brian D. Bue}{jpl}
\icmlauthor{David R. Thompson}{jpl}
\icmlauthor{Vijay Natraj}{jpl}
\icmlauthor{Mario Parente}{umass}
\end{icmlauthorlist}

\icmlaffiliation{cmuri}{The Robotics Institute, Carnegie Mellon University, Pittsburgh, PA, USA}
\icmlaffiliation{jpl}{Jet Propulsion Laboratory, California Institute of Technology, Pasadena, CA, USA}
\icmlaffiliation{umass}{University of Massachusetts, Amherst, MA, USA}

\icmlcorrespondingauthor{Shubhankar Deshpande}{shubhand@cs.cmu.edu}

\icmlkeywords{Machine Learning, Imaging, Earth Science, Climate Change}

\vskip 0.3in
]



\printAffiliationsAndNotice{}  

\begin{abstract}
According to a recent investigation, an estimated 33-50\% of the world's coral reefs have undergone degradation, believed to be as a result of climate change \cite{ISRS}. A strong driver of climate change and the subsequent environmental impact are greenhouse gases such as methane. However, the exact relation climate change has to the environmental condition cannot be easily established. Remote sensing methods are increasingly being used to quantify and draw connections between rapidly changing climatic conditions and environmental impact. A crucial part of this analysis is processing spectroscopy data using radiative transfer models (RTMs) which is a computationally expensive process, and limits their use with high volume imaging spectrometers. This work presents an algorithm that can efficiently emulate RTMs using neural networks leading to a multifold speedup in processing time, and yielding multiple downstream benefits.
\end{abstract}

\section{Introduction}
\label{submission}
Acquatic ecosystems such as coral reefs, kelp beds, seagrass and wetlands serve key roles to sustain the coastal marine populations. They serve as major sites of nutrient cycling, transport, and carbon storage. They occupy only 0.1\% area of the ocean, but they are massive reservoirs of biodiversity \cite{Roberts1280, Bruckner2002} and provide shelter to and support over 25\% of all marine species on the planet \cite{wwf}.
However the balance within these coastal ecosystems is fragile, and is being increasingly threatened due to climate change \cite{10.2307/24860867, nicholls2007coastal} Among these include dangers such as ocean acidification, industrial runoff, and overfishing \cite{Hoegh-Guldberg1737}. While these stressors are evident while analyzing effects on a local scale, the data is insufficient to characterize global scale effects. A large contributor to these stressors is Methane, which is a powerful greenhouse gas that is strongly linked with other trace gases as the cause of global warming, and is the focus of numerous air quality and public health policies \cite{MethaneSurvey}. Many current methane monitoring methods are limited to regional or coarser scale resolution and often cannot detect individual sources. This is a gap that remote sensing\footnote{The term remote sensing is used interchangeably with imaging spectroscopy henceforth.} based techniques have successfully been able to bridge \cite{HOCHBERG2003159}.

The field of imaging spectroscopy itself has a long history that can be traced back to early instruments built at NASA's Jet Propulsion Laboratory in the early 1980s \cite{GAO2009S17}. These instruments were built to measure the solar radiance after reflection from the earth's surface. These measurements can then be analyzed to recover valuable atmospheric and surface properties. Recently, there has been a growing interest to use these instruments to assess the environmental effects of climate change including:
\begin{itemize}
    \item \textit{Oceanic effects:} Monitor ocean circulation, measure the ocean temperature and wave heights, and track sea ice \cite{noaa2009}.
    \item \textit{Land effects:} Track changes in land vegetation, deforestation, and examine the health of indigenous plants and crops \cite{Guo2013}.
    \item \textit{Atmospheric pollution:} Quantify worldwide emissions, concentrations of air pollutants, and their trends. \cite{NAP24938}.
\end{itemize}

Though these are examples scenarios in which imaging spectrometers have successfully been used till date, the near future presents an unprecedented opportunity for rapid progress in applications of remote sensing to benefit society, due to recent computing advances. \cite{NAP24938}.

However, the process of recovering surface properties from imaging data collected by spectrometers is fraught with challenges. Of crucial concern among these is the accurate removal of atmospheric effects from the radiation incident on the sensor. These effects are introduced since the solar radiation on the Sun-surface-sensor path is subject to atmospheric effects such as scattering and absorption. Recovering the true surface parameters requires inverting the measurement with a physical model of the atmospheric radiative transfer in a process called atmospheric correction.

A fundamental component of this process is Radiative Transfer Models (RTMs) \cite{stamnes_thomas_stamnes_2017}, which are responsible for solving the general equations of radiative transport based on known input parameters (such as solar radiance and surface characteristics). This is a computationally expensive process which has to be potentially be run on millions of spectra acquired by high bandwidth imaging spectrometers. Since these instruments produce too high a quantity of data to be processed with an RTM, typically RTMs are used to calculate lookup tables based on the atmospheric conditions observed during image acquisition. These lookup tables can then be used for interpolation during runtime instead of running the full forward RTM computation. Though this approach works in practice, we can only precompute values for a limited number of parameters, since the number of samples needed to represent the state space grows exponentially with the number of input variables. An alternative solution would be to speed up RTMs such that they could be efficiently used at runtime. This would yield multiple downstream benefits during analysis, including enabling the remote sensing community to pursue new surface retrieval and optimal estimation (OE) approaches.

\section{Learning Radiative Transfer Models}
Recent work suggests using non parametric function estimation through models such as Gaussian processes \cite{Martino2017AutomaticEA}, or neural networks \cite{Verrelst2016EmulationOL, Thompson2018a, amt-2018-436} for emulating RTMs. The training dataset would consist of data generated through RTMs over the required range of surface and atmospheric parameters. A function approximator trained on this datd could then act as the RTM for the process of inversion, without having to actually solve the underlying differential equation. To better model the dataset, we designed a neural network architecture that better models the underlying physical parameter space. This section aims to introduce the core ideas behind the RTM emulation technique we designed.

\subsection{Decomposition of parameter space}
The radiative transfer function $F(x)$ can be analytically decomposed into intermediate quantities that are easier to model, which include: solar illumination components, and top of atmosphere reflectance (TOA) $\rho_{obs} = \frac{\mathbf{y}\pi}{\phi_{o}\mathbf{e}_{o}}$, which is normalized for the solar illumination. The quantities $\phi_{o}$ and $\mathbf{e}_{o}$ can then be used to infer the corresponding radiance.
\begin{figure}[hbtp!]
    \centering
    \includegraphics[width=\linewidth]{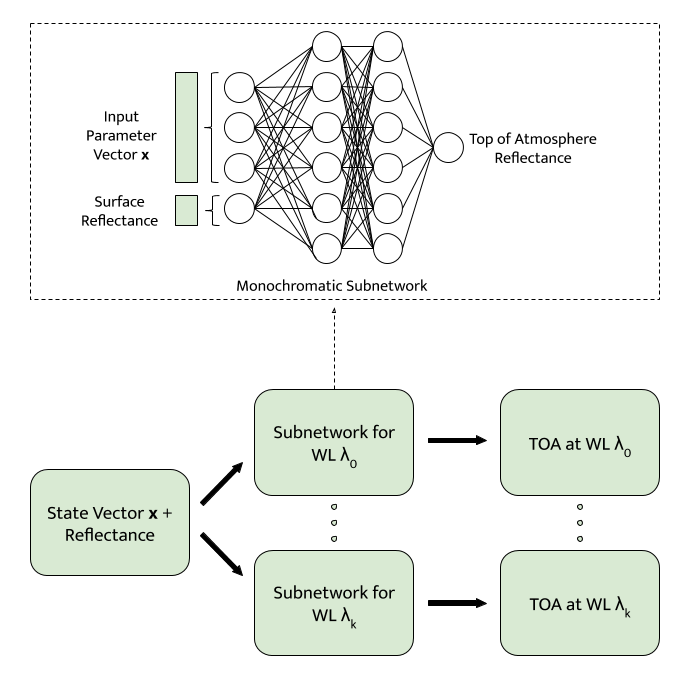}
    \vspace{-1em}
    \caption{Neural RTM}
\end{figure}
\subsection{Monochromatic subnetworks}
The observed radiance within any wavelength channel can be fully specified channel specific the parameters. Formally, absent any prior distribution that couples the influence of neighboring wavelengths, the radiance across channels become conditionally independent of each other given the channel parameters. This insight permits us to decompose the RTM $F(x)$ into intermediate functions $f_i(x)_{i=1}^{k}$ for $k$ different wavelength channels. Using this insight, we can train subnetworks that regress on radiance values in a single channel, instead of regressing across the entire wavelength spectrum. Another benefit of this approach is that the partial derivatives of the radiance channels with respect to their surface reflectance are independent of each other, which simplifies analytical Jacobians calculations during iterative gradient descent inversions \cite{THOMPSON2018355}.

\subsection{Weight Propagation}
The spectral responses for adjacent wavelengths are often correlated, and hence we use the converged weights from subnetworks of previous channels as priors for training the current subnetwork. These weights can then be fine-tuned to estimate the radiances for the current channel. In comparison to training subnetworks from scratch, this approach yields a substantial reduction in training time along with improved accuracy.

\section{Experiments}
We investigate the performance of the neural RTM using data collected by the PRISM \cite{Mouroulis:08, Mouroulis:14} imaging spectrometer. The state space consists of the surface reflectance $\mathbf{\rho_s}$ and the state vector $\mathbf{x}$ containing other parameters relevant to retrieval. A set of values were identified for the input state vector that correspond to typical measurement conditions. For ground truth training data, we generated RTM output using libRadtran, with input parameters relevant to the experiment.
\begin{algorithm}[h]
    \KwInput{$n x m$ matrix $\mathbf{X}$ of $n$ state vectors and $m$ parameters; $n x k$ matrix $\mathbf{Y}$ of k-dimensional $\rho_{obs}$ spectra associated with each state vector at wavelengths $\lambda = {\lambda_i}_{i=1}^{k}$; convergence tolerance $\epsilon$; maximum number of training epochs $\eta_{epoch}$}
    \KwOutput{Neural RTM model $F(\mathbf{x})$ $\rightarrow \mathbf{y}$ consisting of $k$ trained neural networks $f_i$ each mapping $m$ dimensional state vector $\mathbf{x}$ to corresponding $\mathbf{\rho_{obs}}$ response $\mathbf{y}_i$ at wavelength $\lambda_i$}
    \For{$i=1$ to $k$}
    {
        Let $\mathbf{y}_i = \mathbf{Y}$ be the $\rho_{obs}$ responses at wavelength $\lambda_i$ associated with the n state vectors \\
        Partition $(\mathbf{X},\mathbf{y}_i)$ into training, validation, and test sets \\
        Let $f_i$ be an $L$-layer neural network model with a set of weight matrices and bias vectors \\
        \If{$i=1$}
        {
            Create new model for first channel ($r_i$ by initializing weight matrices $W_i$ and bias vectors $b_i$
        }
        \Else
        {
            Propagate weight matrices and bias vectors from the previous model $r_{i-1}$ to current model $f_i$ (i.e. $W_i = W_{i-1}$ and $b_i = b_{i-1}$
        }
        \For{$e=1$ to $\eta_{epoch}$}
        {
            Train current network model $f_i$ on the training set, to minimize the $\rho_{obs}$ prediction error for channel centered at wavelength $\lambda_i$.
            Compute the average error $e_{test}$ applying $f_i$ to the test set
        }
        \If{$e_{test}$ has converged or $e = \eta_{epoch}$}
        {
            \Return Trained model $f_i$
        }
    }
    \Return Trained neural RTM $F(\mathbf{x})$ $\rightarrow \mathbf{y} = {f_i\{\mathbf{x} \rightarrow \mathbf{y}_i\}}_{i=1}^{k}$
\caption{Neural RTM Emulation}
\end{algorithm}
\vspace{-3mm}

\textit{Algorithm 1} describes the procedure used to train the neural RTM. We use a standard feed forward architecture with two hidden layers, and ReLU \cite{Nair2010RectifiedLU} activation. The subnetwork weights corresponding to the first channel were initialized \cite{Glorot2010UnderstandingTD} and the training of subsequent subnetworks used weight propagation. Subnetworks were optimized \cite{Kingma2015AdamAM}, and trained until the mean absolute error converged to with 0.1\%, or the maximum number of epochs $n_{epoch}$ were reached.
\textit{Figure 2}, shows a plot of the $\rho_{obs}$ predicted using the neural RTM vs the standard radiative transfer code (libRadtran), which matches to within 0.1\% mean absolute error.
The code used to train the neural RTM has been released \footnote{\url{https://github.com/dsmbgu8/isofit/}}.
\begin{figure}[hbtp!]
    \centering
    \includegraphics[width=\linewidth]{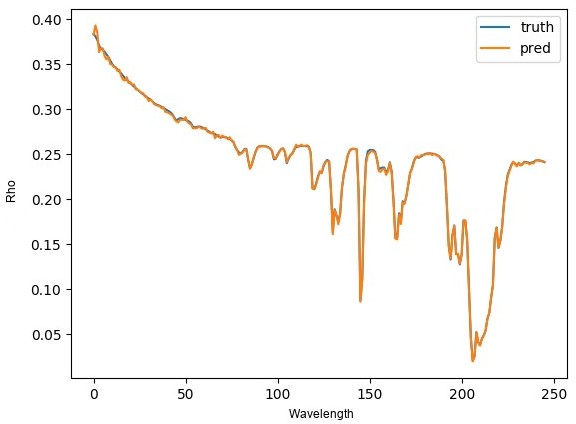}
    \vspace{-1em}
    \caption{Ground truth and predicted $\mathbf{\rho}_{obs}$ spectra}
    \label{fig:my_label}
\end{figure}

\section{Discussion}

We have shown that it is possible to use non parametric function estimation models for the task of emulating radiative transfer. In the process we introduced a network architecture and training techniques that help to better model the underlying physical parameter space. This formulation offers a path to increase the interpolation accuracy of current approaches, reduce the runtime by several orders of magnitude, and will enable the imaging community to pursue new surface retrieval approaches. More broadly, the field of remote sensing and earth science are ripe for impact due to the exponential growth rate in data generation coupled with recent developments in machine learning. We have presented one such example of research that will have potential impact in enabling the earth science community to understand, quantify and draw connections between rapidly changing climatic conditions and environmental impact.

\section*{Acknowledgements}

We thank Michael Eastwood and Robert O. Green for providing expertise in the theory and application of imaging spectroscopy and atmospheric correction methods. We further thank Terry Mullen for contributing to the analysis and generating RTM outputs.  This work was undertaken in part at the Jet Propulsion Laboratory, California Institute of Technology, under contract with NASA.


\nocite{THOMPSON201718}
\nocite{gmd-9-1647-2016}

\bibliography{example_paper}
\bibliographystyle{icml2019}

\end{document}